\documentclass{article}
\usepackage{spconf,amsmath,graphicx}
\usepackage{adjustbox,booktabs,multirow,multicol,xcolor}
\usepackage{tikz,enumitem,titlesec}
\usepackage{comment}
\definecolor{DarkRed}{rgb}{0.7, 0.0, 0.0}
\definecolor{DarkGreen}{rgb}{0.0, 0.3, 0.0}

\title{Real-time Traffic Accident Anticipation with Feature Reuse}
\name{Inpyo Song$^{\star}$, Jangwon Lee$^{\dagger}$\thanks{$^{\star}$ songinpyo@skku.edu, $^{\dagger}$ Corresponding author: leejang@skku.edu}}
\address{Department of Immersive Media Engineering, Sungkyunkwan University, South Korea}

\begin{document}
%\ninept
%
\maketitle
\begin{abstract}
This paper addresses the problem of anticipating traffic accidents, which aims to forecast potential accidents before they happen.
%Traffic accidents remain a pressing global concern despite the rapid evolution of autonomous driving technologies.
Real-time anticipation is crucial for safe autonomous driving,
yet most methods rely on computationally heavy modules like optical flow and intermediate feature extractors,
making real-world deployment challenging.
%Anticipating accidents in real time is crucial for proactive decision-making and robust driving safety,
%yet most existing methods rely on computationally heavy modules such as optical flow, object tracking,
%and image/video feature extractors that challenge deployment under real-world resource constraints.
In this paper, we thus introduce \textbf{RARE} (Real-time Accident anticipation with Reused Embeddings),
a lightweight framework that capitalizes on intermediate features from a single pre-trained object detector.
By eliminating additional feature-extraction pipelines, RARE significantly reduces latency.
Furthermore, we introduce a novel \emph{Attention Score Ranking Loss},
which prioritizes higher attention on accident-related objects over non-relevant ones.
%In addition, we also introduce an Attention Score Ranking Loss that enforces higher attention on accident-related objects than on non-relevant ones.
This loss enhances both accuracy and interpretability.
RARE demonstrates a 4-8\,$\times$ speedup over existing approaches on the DAD and CCD benchmarks, achieving a latency of 13.6\,$ms$ per frame (73.3\,FPS) on an RTX 6000.
Moreover, despite its reduced complexity,
it attains state-of-the-art Average Precision and reliably anticipates imminent collisions in real time.
These results highlight RARE’s potential for safety-critical applications where timely and explainable anticipation is essential.
\end{abstract}

\section{Introduction} \label{sec1:introduction}
%%%%%%%%%% 1. Big Problem %%%%%%%%%%%%
%Autonomous driving technology has witnessed rapid growth, with advances in both fully autonomous vehicles and Advanced Driver Assistance Systems (ADAS).
%Despite these developments, road traffic accidents remain a critical global concern. According to the World Health Organization, approximately 1.35 million fatalities and up to 50 million non-fatal injuries result from road accidents each year \cite{world2019global}.
Autonomous driving technology has advanced rapidly, with significant progress in both fully autonomous vehicles and Advanced Driver Assistance Systems (ADAS) \cite{chen2022milestones}.
However, road traffic accidents remain a major global concern.
According to the World Health Organization, approximately 1.35 million fatalities
and up to 50 million non-fatal injuries result from road accidents each year \cite{world2019global}.
%Furthermore, recent statistics show that autonomous vehicles can experience a crash rate of 9.1 per million miles, compared to 4.1 per million miles for conventional cars \cite{chougule2023comprehensive}.
Furthermore, alarmingly, recent statistics show that autonomous vehicles experience a crash rate of 9.1 per million miles
—higher than the 4.1 per million miles for conventional cars \cite{chougule2023comprehensive}.
%These statistics underscore the urgent need for intelligent accident anticipation systems
%to minimize imminent risks and enhance overall driving safety.
These statistics highlight the critical need for real-time intelligent accident anticipation systems
that can actively mitigate risks and improve overall driving safety.
% Addressing this challenge is essential to ensure the safe integration of autonomous vehicles into everyday traffic systems.
%
%%%%%%%%%% 2. Other works tried to solve problem %%%%%%%%%%%%
%Many existing Traffic Accident Anticipation (TAA) methods rely on computationally intensive modules such as optical flow, object tracking, and feature extraction networks to capture complex spatio-temporal dynamics \cite{DAD_chan2017anticipating,DSTA_karim2022dynamic,AM-Net_ROL_karim2023attention,GSC_wang2023gsc,UString_CCD_bao2020uncertainty}.
Therefore, extensive research has been carried out on Traffic Accident Anticipation (TAA) \cite{DAD_chan2017anticipating,UString_CCD_bao2020uncertainty}.
However, such approaches often rely on computationally intensive modules,
such as optical flow, object tracking, and feature extraction networks, to capture complex spatio-temporal dynamics
\cite{AM-Net_ROL_karim2023attention,GSC_wang2023gsc,DSTA_karim2022dynamic}.
While these techniques often yield high accuracy, they impose substantial computational overhead and lengthy inference times,
making real-world deployment challenging.
%impeding real-time deployment in resource-constrained environments.
%As illustrated in Figure~\ref{Fig1:APmTTA_chart}, the disparity in latency across different methods underscores the urgency for lightweight solutions that preserve strong predictive performance.

\begin{figure}[!t]
\includegraphics[width=\columnwidth]{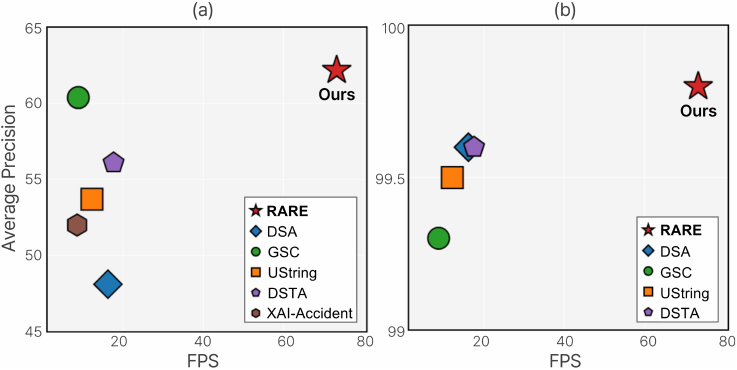}
\caption{
\textbf{Performance comparison of traffic accident anticipation methods}.
%\textbf{Performance and efficiency comparison of traffic accident anticipation methods}.
It shows FPS (x-axis) vs. Average Precision (y-axis) on the DAD (a) and CCD (b) datasets.
RARE is the only method that achieves real-time performance (73.3\,FPS),
exceeding the 30 FPS threshold while maintaining the highest AP on both datasets.
%RARE, is the only method achieving real-time performance (73.3\,FPS),
%surpassing the real-time threshold (30\,FPS) while maintaining the best AP on both datasets.
}
\vspace{-1em}
\label{Fig1:APmTTA_chart}
\end{figure}

%%%%%%%%%% 3. Our contribution %%%%%%%%%%%%
To address these limitations, we introduce \textbf{RARE} (Real-time Accident anticipation with Reused Embeddings),
a streamlined framework that balances inference speed, accuracy, and interpretability.
Unlike prior pipelines that incorporate multiple sub-networks, RARE reuses intermediate features from a single pre-trained object detector, eliminating the need for standalone feature-extraction branches.
This design drastically reduces complexity and improves efficiency.
Figure~\ref{Fig1:APmTTA_chart} clearly shows that the proposed RARE is the only approach that can achieve real-time performance
(73.3\,FPS on an RTX 6000) while maintaining the highest accuracy compared to previous approaches.
Additionally, we present an \textbf{Attention Score Ranking Loss} that explicitly supervises object-level attention, ensuring the model consistently focuses on accident-related objects rather than relying solely on implicit attention learning.

%%%%%%%%%% 4. Experiments %%%%%%%%%%%%
We evaluated our method on two widely used public TAA benchmarks, DAD and CCD \cite{DAD_chan2017anticipating,UString_CCD_bao2020uncertainty}.
RARE achieves a notable 4–8,$\times$ faster inference time compared to established baselines, maintaining state-of-the-art Average Precision.
Such performance, combined with attention-based interpretability, demonstrates RARE’s practical viability in safety-critical applications where timely and transparent accident anticipation is paramount.

%%%%%%%%%% 5. Summary %%%%%%%%%%%%
In summary, our key contributions are:
\vspace{-0.5em}
\begin{itemize}[leftmargin=1em]
\setlength\itemsep{0em}
\item \textbf{Efficient Framework}: We introduce RARE, a real-time TAA framework that reuses embeddings from a single object detector, eliminating heavy optical flow or feature extraction networks.
\item \textbf{Attention Ranking Loss}: We propose a loss function that enforces higher attention on accident-critical objects, boosting both interpretability and predictive consistency.
\item \textbf{Significant Speed-up}: RARE achieves best or near-best performance in AP and a 4–8\,$\times$ speed advantage on two benchmarks.
\end{itemize}

\section{Related Work} \label{sec2:related work}
Traffic Accident Anticipation (TAA) has emerged as a critical research area, fueled by the need to foresee collisions before they occur in both autonomous driving and advanced driver assistance contexts.
Early TAA methods typically rely on image-based cues, extracting spatial and object-level features from standard backbones (e.g., VGG \cite{vgg_simonyan2014very}) or pre-trained detectors \cite{ren2016fasterrcnn}, then feeding these features into spatio-temporal modeling modules \cite{DAD_chan2017anticipating,adalea_suzuki2018anticipating}.

Recently, a popular trend in TAA is the integration of attention mechanisms to capture intricate interactions among scene elements.
For instance, DSTA \cite{DSTA_karim2022dynamic} employs attention modules to fuse scene-level and object-level features, facilitating more robust spatio-temporal reasoning.
Similarly, AM-Net \cite{AM-Net_ROL_karim2023attention} integrates optical flow and object tracking with attention-based feature fusion to highlight critical regions in an evolving traffic scenario.
Graph-based approaches extend this idea, representing vehicles or pedestrians as nodes within a graph neural network.
GSC \cite{GSC_wang2023gsc} and UString \cite{UString_CCD_bao2020uncertainty} use graph-based reasoning to model agent interactions, improving the detection of collision-prone situations.
Beyond static appearance cues, several methods emphasize predicting future trajectories of traffic participants.
FOL \cite{FOL_A3D_yao2019unsupervised,DoTA_yao2022dota} employed a framework that forecasts the future location of each object to identify potential collisions.

%Given the safety-critical nature of the prediction of traffic accidents, the ability to explain has gained increasing attention.
Another significant research direction, driven by the safety-critical nature of traffic accident prediction, focuses on the explainability of these predictions. 
Various works employ Grad-CAM or saliency methods \cite{bao2021drive,XAI_karim2022toward} to identify critical regions associated with imminent accidents.
In addition, more recent research \cite{DRAMA_malla2023drama,MM-AU_fang2024abductive} integrates multi-modal cues for deeper interpretability, aiming to uncover not just where, but also why an accident might occur.
Such transparent predictions are particularly valuable for building trust in autonomous driving systems.

While these methods have advanced TAA accuracy and interpretability, many still depend on multiple heavy components (e.g., optical flow, object tracking, or separate feature-extraction networks), hindering real-time deployment.
In contrast, our RARE framework reuses intermediate embeddings from a single pre-trained detector, drastically reducing the computational load.
Furthermore, although multiple TAA methods leverage attention mechanisms \cite{AM-Net_ROL_karim2023attention,DSTA_karim2022dynamic}, they typically learn attention weights implicitly from binary accident labels. 
Our Attention Score Ranking Loss imposes explicit constraints on object attention, ensuring the network consistently prioritizes accident-relevant objects.

\begin{figure*}[!t]
\centering
\includegraphics[width=\linewidth]{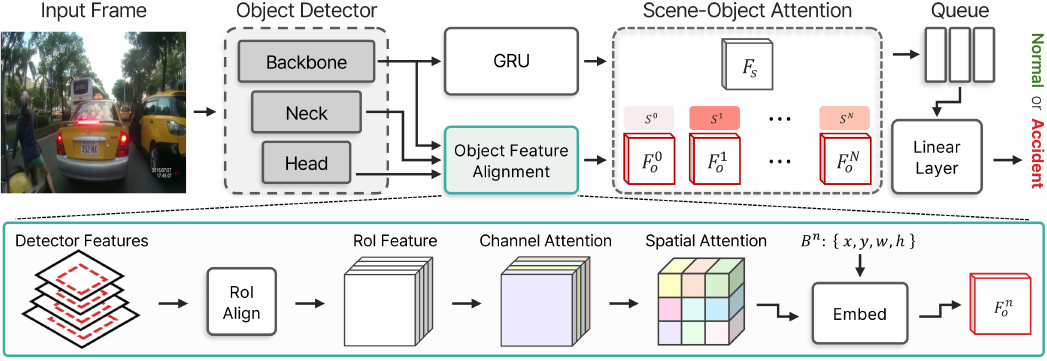}
\caption{
\textbf{Overview of our proposed RARE framework}.
RARE detects objects in each input frame and extracts intermediate features using a pre-trained object detector.
Object-specific embeddings $F_o$ are computed via RoI Align on multi-scale features, while temporal scene-level dynamics are encoded by a GRU over the detector’s backbone features.
Attention module fuses these scene-level and object-level representations, and the fused features are passed to a classifier for predicting accident risk.
}
\label{Fig2:architecture}
\vspace*{-0.5em}
\end{figure*}

\section{Proposed Approach} \label{sec3:proposed approach}
\begin{comment}
We present \textbf{RARE} (\textbf{R}eal-time \textbf{A}ccident anticipation with \textbf{R}eused \textbf{E}mbeddings), a framework designed to deliver accurate accident anticipation under real-time constraints.
Unlike methods that use multiple feature-extraction modules or optical flow, RARE repurposes intermediate embeddings from a single pre-trained object detector, sharply reducing computational overhead.
\end{comment}
In this section, we present \textbf{RARE} (\textbf{R}eal-time \textbf{A}ccident anticipation with \textbf{R}eused \textbf{E}mbeddings),
a framework that delivers accurate accident anticipation under real-time constraints by recycling intermediate embeddings from a single pre-trained object detector.

\subsection{Preliminary Setup}
Let a video sequence be denoted by $V = \{I_1, I_2, \dots, I_{T}\}$, where $I_t$ represents the $t$-th frame.
We use a YOLO object detector \cite{wang2024yolov10} to obtain bounding boxes $B = \{B^1, B^2, \dots, B^N\}$ for the  N  objects in each frame.
Additionally, we extract intermediate feature maps $F_{\text{backbone}}$ and $F_{\text{neck}}$ from the detector.
% We use a YOLO object detector \cite{wang2024yolov10} to obtain:
% \begin{itemize}[leftmargin=1em]
%     \item \textbf{Bounding Boxes:} $B = \{B^1, B^2, \dots, B^N\}$ for the $N$ objects in each frame.
%     \item \textbf{Intermediate Feature Maps:} $F_{\text{backbone}}$ and $F_{\text{neck}}$ from the detector, which serve as the core features reused by RARE.
% \end{itemize}

To capture scene-level temporal context, we feed $F_{\text{backbone}}$ from consecutive frames into a Gated Recurrent Unit (GRU):
\begin{equation}
F_{\text{s}} = \text{GRU}\bigl(F_{\text{backbone}}\bigr)
\end{equation}
yielding a hidden state $F_{\text{s}}$ that encodes evolving global scene information.

% Formally, let a video sequence be denoted by $V = I_1, I_2, \dots, I_{T}$ where $I_t$ is the $t$-th frame.
% In each frame $I_t$ the object detector provides both bounding boxes and feature maps.
% These form the backbone of our spatio-temporal modeling for accident anticipation.

% \subsection{Object Detector}
% We employ a YOLO-based object detector \cite{wang2024yolov10} to extract bounding boxes $B = B^1, B^2, \dots, B^N$ and intermediate feature maps $F_{\text{backbone}}$ and $F_{\text{neck}}$.

% Here $N$ is the number of detected objects in frame $I_t$.

% To capture scene-level temporal dynamics over time, we pass the backbone features across frames to a Gated Recurrent Unit (GRU):

% \begin{equation}
% F_{\text{s}} = \text{GRU}(F_{backbone}).
% \end{equation}
% This yields a compact hidden representation $F_{\text{s}}$ that encodes temporal information at the scene level.

\subsection{Object Feature Alignment}
While $F_{\text{s}}$ captures overall scene dynamics, we also need localized object information.
We use RoI Align \cite{maskrcnn_he2017mask} on the multi-scale features $F_{\text{backbone}}$ and $F_{\text{neck}}$, to extract object-centric embeddings:
\begin{equation}
F_{\text{RoI}}^n = \text{RoIAlign}\bigl(B^n, F_{\text{backbone}}, F_{\text{neck}}\bigr), 
\quad n = 1 \dots N.
\end{equation}

Each $F_{\text{RoI}}^n$ captures localized appearance cues for the $n$-th bounding box.
We then apply a Convolutional Block Attention Module (CBAM) \cite{woo2018cbam} to enhance these embeddings by leveraging channel and spatial attention.
Additionally, bounding box coordinates $B^n$ are embedded and concatenated with the RoI features.
Finally, a set of linear layers fuses these components into a single object-specific embedding:
\begin{equation}
F_{\text{o}}^n = \text{Embed}(\text{CBAM}(F_{\text{RoI}}^n), B^n)
\end{equation}
This operation yields a robust feature vector $F^n_{\text{o}}$ that represents both the local visual cues (via RoI features) and the positional context (via bounding box coordinates).

\subsection{Scene-Object Attention}
To fuse global scene context with local object features, we adopt a multi-head attention (MHA) mechanism:
\begin{equation}
F_{f}, S^n 
= \text{MHA}(F_{\text{s}}, \{F_{\text{o}}^1, \dots, F_{\text{o}}^N\})
\end{equation}
where $F_{f}$ is a fused feature that captures interactions between scene-level and object-level information.
Simultaneously, $S^n$ denotes the attention score for each object, providing interpretability regarding which objects most influence the accident prediction.

\paragraph{Attention Score Ranking Loss}
Typically, existing TAA methods implicitly learn attention based on accident labels.
In contrast, we incorporate an explicit ranking loss to ensure that object bounding boxes covering accident-related regions consistently receive higher attention scores than non-accident boxes.
Specifically, we assign each box $B^n$ as positive if it overlaps ground-truth accident-related bounding boxes based on Intersection over Union (IoU) value over 0.5, or negative otherwise.
Let $S_{a}$ and $S_{na}$ be the attention scores for positive and negative boxes, respectively.
Our ranking loss $L_R$ is defined as:
\begin{equation}
L_R = 
\min (0,\; \max(S_{na}) + m - \min(S_a))
\end{equation}
where $m$ represents the margin.
The loss penalizes cases where any non-accident box receives a higher attention score than any accident-related box, ensuring a clear distinction between these two objects.

\begin{table*}[t]
    \centering
    \setlength{\tabcolsep}{8pt}
    \begin{tabular}{l | c c | c c c c }
    \toprule
    \multirow{2}{*}{Model} & \multirow{2}{*}{FPS $\uparrow$}& \multirow{2}{*}{Latency(ms) $\downarrow$} & \multicolumn{2}{c}{DAD} & \multicolumn{2}{c}{CCD} \\
    \cline{4-7}
    % \cmidrule{3-8}
    & & & AP(\%) $\uparrow$ & mTTA(s) $\uparrow$ & AP(\%) $\uparrow$ & mTTA(s) $\uparrow$ \\
    \midrule
    DSA \cite{DAD_chan2017anticipating} & 16.2 & 61.7 & 48.1 & 1.34  & \underline{99.6} & 4.52 \\
    L-RAI \cite{L-RAI_zeng2017agent} & - & - & 51.4 & 3.01 & - & - \\
    AdaLEA \cite{adalea_suzuki2018anticipating} & - & - & 52.3 & 3.43 & - & - \\
    UString \cite{UString_CCD_bao2020uncertainty} & 12.2 & 82.2 & 53.7 & 3.53  & 99.5 & \underline{4.74} \\
    DSTA \cite{DSTA_karim2022dynamic} & 17.6 & 56.7 & 56.1 & \underline{3.66}  & \underline{99.6} & \textbf{4.87} \\ % 41.7 in paper
    XAI-Accident \cite{XAI_karim2022toward} & *8.5 & *117.7  & 52.0 & \textbf{3.79} & 94.0 & 4.57 \\
    GSC \cite{GSC_wang2023gsc} & *8.8 & *113.8 & \underline{60.4} & 2.55  & 99.3 & 3.58 \\
    \midrule
    \multicolumn{7}{c}{\textit{Real-time Processing} ( Latency $ < 30ms$ )} \\
    \midrule
    % RARE (Ours) & \textbf{73.3} & \textbf{13.6} & \underline{58.3} & 2.58 & \textbf{99.8} & 4.03 \\
    RARE (Ours) & \textbf{73.3} & \textbf{13.6} & \textbf{62.2} & 2.51 & \textbf{99.8} & 4.03 \\
    \bottomrule
    \end{tabular}
    \caption{
    Comparison of Traffic Accident Anticipation methods, highlighting the trade-off between accuracy (AP), earliness (mTTA), and inference speed (FPS/Latency).
    * indicates that the inference speed was reported under different experimental conditions than ours and was directly taken from the original paper.
    \textbf{Bold} denotes the best performance, while \underline{underline} represents the second-best.
    }
    \label{Table:Quantitative Results}
    \vspace*{-1em}
\end{table*}

\subsection{Accident Anticipation}
Finally, to incorporate temporal context over successive frames, we maintain a queue of size $k$ containing the most recent fused features:
\begin{equation}
\text{Queue}^t = [F_f^t,\; F_f^{t-1},\ldots,F_f^{\,t-k+1}]
\end{equation}

This queue serves as a short-term memory, capturing how accident cues evolve over consecutive frames.
We pass  $\text{Queue}_t$ through a small classifier (two fully connected layers) to obtain the accident risk score $\hat{y}_t$:
\begin{equation}
\hat{y}_t = \text{Classifier}(\text{Queue}_t)
\end{equation}

Hence, RARE leverages both spatial (object and scene) and temporal (recent frames) information to predict whether an accident is imminent.

\subsection{Training Objective}
To encourage earlier accident predictions, we adopt the AdaLEA loss \cite{adalea_suzuki2018anticipating}, which dynamically adjusts per-frame penalty weights based on both the distance to the accident onset and the model’s evolving ability to anticipate collisions.
It increases penalties for frames occurring earlier than the model’s current average anticipation time, progressively shifting focus from near-accident frames to earlier ones.
Negative samples follow standard binary cross-entropy.
Our final objective combines AdaLEA loss with Attention Score Ranking Loss.
\begin{equation}
    L = L_{\mathrm{AdaLEA}} + \gamma \, L_{R}
\end{equation}
where $\gamma$ is a loss weighting hyperparameter.

% We optimize a {combined} loss function:
% \begin{equation}
% L = L_F + \gamma L_R,
% \end{equation}
% where $L_R$ is the attention score ranking loss described above, and $\gamma$ is a balancing hyperparameter.

% \paragraph{Frame-Wise Exponential Loss}
% For $L_F$, we adopt the frame-wise exponential loss from \cite{DAD_chan2017anticipating}, which prioritizes early correctness by assigning larger penalties as the accident moment approaches.
% Formally, if $l_V \in \{0,1\}$ indicates whether video $V$ contains an accident and $\tau$ denotes the accident onset frame,
% \begin{align}
%     L_F \;=\;& 
%     -\,l_V \sum_{t=1}^T 
%     e^{-\max\bigl(\frac{\tau - t}{f},\,0\bigr)} \log \hat{y}_{t} \\
%     & -\,(1 - l_V)\sum_{t=1}^T \log\bigl(1 - \hat{y}_{t}\bigr), \nonumber
% \end{align}
% where $f$ is the frame rate. For frames preceding the accident, the exponential term $e^{-\max\left(\frac{\tau - t}{f}, 0\right)}$ imposes increasingly strong penalties on misclassifications. In non-accident videos ($l_V=0$), the standard binary cross-entropy term applies.

\section{Experimental Results} \label{sec4:experimental results}
\subsection{Implementation Details}
We train RARE using Stochastic Gradient Descent (SGD) with a learning rate of $5\times 10^{-2}$ and a batch size of 4. All input frames are resized to $640\times 640$, and we retain object detections with confidence above $0.1$, focusing on classes \{\emph{person, bicycle, car, motorcycle, bus, truck}\}. 
We empirically set the margin $m=0.1$, loss-balancing hyperparameter $\gamma$ to 10 and temporal queue size $k=10$.
All experiments run on an AMD EPYC 9124 CPU and an NVIDIA RTX 6000 Ada GPU.
% with a batch size of 1 to simulate real-time, online inference scenarios in which frames arrive sequentially.
Since some prior works \cite{adalea_suzuki2018anticipating,XAI_karim2022toward} do not provide public code,
preventing execution in our environment due to hardware differences, we report their published FPS/latency (* in Table~\ref{Table:Quantitative Results}).
%Some prior works \cite{adalea_suzuki2018anticipating,XAI_karim2022toward} do not provide public code, preventing execution in our environment and leading to hardware differences.
%Thus, we report their published FPS/latency (* in Table~\ref{Table:Quantitative Results}).
Additionally, GSC~\cite{GSC_wang2023gsc} excludes preprocessing time (object detection + feature extraction).
So, we measured this overhead separately and added it for a fair end-to-end comparison.
% Some prior works \cite{adalea_suzuki2018anticipating,XAI_karim2022toward} do not publicly provide code or use different hardware, preventing direct speed measurement under our environment.
% Hence, we report their published FPS/latency values (marked with an *) in Table~\ref{Table:Quantitative Results}.
% In addition, GSC~\cite{GSC_wang2023gsc} does not include preprocessing time (object detection + feature extraction) in its latency reports.
% We therefore measured that overhead separately in our setup and added it to GSC’s model-processing time for a fair end-to-end comparison.

\subsection{Datasets}

\noindent \textbf{Dashcam Accident Dataset} (\textbf{DAD}) \cite{DAD_chan2017anticipating} is a widely recognized TAA benchmark comprising 620 positive accident videos and 1,130 negative videos, each 5 seconds long at 20\,FPS (100 frames per video). 
Accidents occur within the final 0.5 seconds of positive videos, while negative videos contain no accidents. 
Following the standard split, we use 455 positive and 829 negative videos for training, and 165 positive and 301 negative for testing.

\noindent \textbf{Car Crash Dataset} (\textbf{CCD}) \cite{UString_CCD_bao2020uncertainty} also serves as a standard benchmark for TAA, comprising 4,500 dashcam videos at 10 fps, each spanning 5 seconds (50 frames).
Accidents occur within the last 2 seconds in positive clips. 
CCD is split 80\,\% / 20\,\% for training and testing, respectively, encompassing diverse environments (weather, traffic density).

\subsection{Evaluation Metrics}
We adopt two main metrics to assess correctness, and earliness following standard practices in TAA research \cite{DAD_chan2017anticipating,UString_CCD_bao2020uncertainty}.

Average Precision (AP) measures the area under the precision-recall curve, capturing the trade-off between precision and recall.
It is particularly informative in imbalanced datasets, emphasizing the ability to detect true positives.

Time-to-Accident (TTA) evaluates how early the model predicts an accident relative to its occurrence.
% $\text{TTA} = \max \{\tau - t | P(t) \geq \theta, \, 1 \leq t \leq \tau\}$
% where  $\tau$  denotes the ground truth accident time,
% $t$ is the frame index, and  $P(t)$  is the model’s predicted probability.
Higher TTA values indicate earlier predictions relative to the actual event.
Mean Time-to-Accident (mTTA) averages TTA values across varying thresholds, providing a threshold-agnostic view of earliness.

% We report the processing time per frame (in milliseconds) to evaluate how efficiently the model operates in a streaming context.
% Real-time constraints typically demand $Latency < 30\,ms$ per frame.

\subsection{Comparison with Baselines}
We compare RARE against seven existing methods, focusing on both prediction quality (AP, mTTA) and inference speed (FPS, latency).
Table~\ref{Table:Quantitative Results} shows a detailed quantitative comparison.
Notably, RARE (Ours) exhibits the fastest inference with 73.3 FPS, corresponding to a per-frame latency of 13.6 ms—significantly below the real-time threshold of 30 ms.
Despite its 4–8\,$\times$ speed advantage, RARE still achieves 62.2\,\% AP on DAD and 99.8\,\% AP on CCD, maintaining performance exceeding prior state-of-the-art.
Overall, these results validate RARE’s capability to detect potential accidents at a frame rate suitable for real-time applications, all while preserving solid AP scores.
We attribute this strong efficiency and performance by reusing detector embeddings and attention score ranking loss.

\begin{table}[t]
    \centering
    \begin{adjustbox}{max width=\columnwidth}
    \begin{tabular}{l | c c | c c}
    \toprule
    \multirow{2}{*}{Model} & \multirow{2}{*}{FPS $\uparrow$}& \multirow{2}{*}{Latency(ms) $\downarrow$} & \multicolumn{2}{c}{DAD}\\
    \cline{4-5}
    % \cmidrule{3-8}
    & & & AP(\%) $\uparrow$ & mTTA(s) $\uparrow$ \\
    \midrule
    UString \cite{UString_CCD_bao2020uncertainty} & 12.2 & 82.2 & 58.6 & \underline{2.12} \\
    DSTA \cite{DSTA_karim2022dynamic} & 17.6 & 56.7 & \underline{59.2} & 2.07  \\
    \midrule
    RARE (Ours) & \textbf{73.3} & \textbf{13.6} & \textbf{62.2} & \textbf{2.51} \\
    \bottomrule
    \end{tabular}
    \end{adjustbox}
    \caption{
    Comparison of TAA methods, highlighting the trade-off between accuracy (AP), earliness (mTTA), and inference speed (FPS/Latency) on DAD dataset.
    }
    \label{Table:Quantitative Results2 focusing on AP}
\end{table}

\begin{table}[t]
    \centering
    \begin{adjustbox}{max width=\columnwidth,max height=4em}
    \begin{tabular}{ l | c c | c c}
        \toprule
        Model Variant & \multicolumn{2}{c|}{AP (\%) $\uparrow$} & \multicolumn{2}{c}{mTTA (s) $\uparrow$} \\
        \midrule
        (a) \;RARE & \textbf{62.2} & (-) & 2.51 & (-) \\
        (b)\quad w/o $\mathcal{L}_R$  & 56.3 & (\textcolor{DarkRed}{-\small{5.9}}) & 2.53 & (\small{+0.02}) \\
        (c)\quad w/o $F_{\text{backbone}}$ & 54.0 & (\textcolor{DarkRed}{-\small{8.2}}) & 2.58 & (\small{+0.07}) \\
        (d)\quad w/o $F_{\text{neck}}$ & 56.0 & (\textcolor{DarkRed}{-\small{6.2}}) & 2.52 & (\small{+0.01})  \\
        \bottomrule
    \end{tabular}
    \end{adjustbox}
    \caption{
    Ablation study on RARE components on DAD dataset.
    Decreased values are highlighted in \textcolor{DarkRed}{red}.
    }
    \label{tab:ablation}
\end{table}

\noindent\textbf{Does achieving high AP compromise mTTA?}
% \paragraph{Does achieving high AP compromise mTTA?}
Several works suggest a trade-off between AP and mTTA \cite{DAD_chan2017anticipating,GSC_wang2023gsc}.
Our findings support this trade-off in general but also highlight that RARE achieves a more balanced performance across both metrics.
As shown in Table~\ref{Table:Quantitative Results2 focusing on AP}, the results reported in \cite{GSC_wang2023gsc} indicate that after further training, both DSTA~\cite{DSTA_karim2022dynamic} and UString~\cite{UString_CCD_bao2020uncertainty} achieve higher AP, approaching 60\,\%.
However, this improvement comes at the cost of a significant mTTA drop, approximately 1.5\,s.
In contrast, RARE not only surpasses 60\,\% AP but also maintains a higher mTTA than these methods, demonstrating a superior balance between detection accuracy and anticipation performance.

\subsection{Ablation Study}
To assess each component’s impact, we remove key modules from RARE and report the DAD performance in Table~\ref{tab:ablation}. 
Removing the \emph{Attention Score Ranking Loss} (b) significantly reduces AP (by 5.9\%) while slightly increasing mTTA, suggesting more false positives are triggered earlier.
Likewise, omitting either the backbone or neck features in the \emph{Object Feature Alignment} step (c, d) substantially lowers AP, indicating that multi-scale embeddings are crucial for robust spatio-temporal reasoning.
These findings verify that our design choices—particularly explicit attention supervision and reusing both backbone/neck features are key to RARE’s reliable and accurate accident anticipation.

\subsection{Qualitative Results}
\begin{figure}[!t]
\centering
\includegraphics[width=\columnwidth]{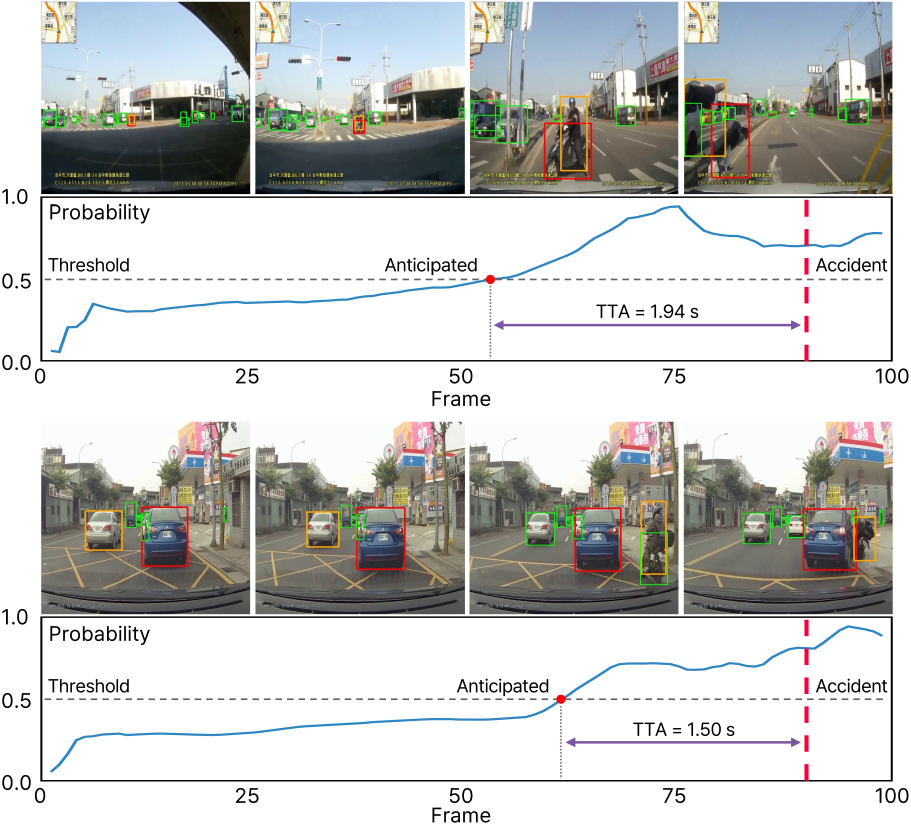}
\caption{
\textbf{Qualitative results of RARE on DAD dataset}.
The top row shows detected bounding boxes in green, with the highest attention-weighted in red and the second in orange.
The bottom row presents the predicted risk score in blue, with the vertical red dotted line marking the accident timing and the horizontal black dotted line indicating the threshold.
}
\label{Fig3:qualitative results}
\end{figure}

Figure~\ref{Fig3:qualitative results} demonstrates RARE’s attention-based interpretability.
The model robustly identifies a high-risk object (red bounding box) long before the accident occurs.
Even when an accident object appears very small or emerges mid-sequence, our model highlights it early, demonstrating robust focus on critical cues.
Meanwhile, the probability curve (bottom plot) rises steadily as the collision approaches, confirming that RARE not only spots the crucial object early but also escalates its risk assessment accordingly.

\section{Conclusion} \label{sec5:conclusion}
In this paper, we presented RARE, a lightweight framework for real-time traffic accident anticipation that addresses the long-standing trade-off between inference speed and predictive performance.
By reusing detector-derived intermediate features, our method eliminates the need for multiple heavy feature extractors, thereby offering real-time inference with a latency of just 13.6 ms per frame.
Moreover, our scene-object attention module with Attention Score Ranking Loss not only effectively fuses local and global cues but also enhances transparency in the decision-making process.
Despite its minimal computational overhead, RARE retains strong predictive capabilities, achieving the highest AP scores on both the DAD and CCD datasets.
%Experimental evaluations showed that RARE took a step toward bridging the gap between accurate accident forecasting and real-time feasibility,
%contributing to the development of safer and more efficient autonomous driving systems.
Experimental evaluations showed that RARE bridges the gap between accurate accident forecasting and real-time feasibility,
contributing to safer and more efficient autonomous driving systems.

\paragraph{Acknowledgement}
% \noindent \textbf{Acknowledgement.}
This research was supported by the MSIT (Ministry of Science and ICT), Korea, under the Graduate School of Metaverse Convergence Support Program (IITP-2025-RS-2023-00254129) and the Global Scholars Invitation Program (RS-2024-00459638), both supervised by the IITP (Institute for Information \& Communications Technology Planning \& Evaluation)

\vfill\pagebreak

\bibliographystyle{IEEEbib}
\bibliography{strings,refs}

\end{document}